\documentclass[runningheads]{llncs}

\usepackage{hyperref}

\usepackage{graphicx}
\usepackage{comment}
\usepackage{amsmath,amssymb} 
\usepackage{color}
\usepackage{bm}
\usepackage{algorithm,algorithmic}
\usepackage{booktabs}
\usepackage{multirow}
\usepackage{epsfig}
\usepackage{amsmath} 
\usepackage{amssymb}
\usepackage{subfigure}
\usepackage{diagbox}

\begin{document}
\pagestyle{headings}
\mainmatter
\def\ECCVSubNumber{2394}
\title{Resolution Switchable Networks for Runtime Efficient Image Recognition}

\titlerunning{Resolution Switchable Networks for Runtime Efficient Image Recognition}

\author{Yikai Wang\inst{1}\thanks{This work was done when Yikai Wang was an intern at Intel Labs China, supervised by Anbang Yao who is responsible for correspondence.} 
\and
Fuchun Sun\inst{1} \and
Duo Li\inst{2} \and Anbang Yao\inst{2}}
\authorrunning{Y. Wang, F. Sun, D. Li and A. Yao}

\institute{Beijing National Research Center for Information Science and Technology$\,$(BNRist),\\ State Key Lab on Intelligent Technology and Systems,\\ Department of Computer Science and Technology, Tsinghua University \and Cognitive Computing Laboratory, Intel Labs China\\
\email{\{wangyk17@mails.,fcsun@\}tsinghua.edu.cn, \{duo.li,anbang.yao\}@intel.com}}

\maketitle

\begin{abstract}
We propose a general method to train a single convolutional neural network which is capable of switching image resolutions at inference. Thus the running speed can be selected to meet various computational resource limits. Networks trained with the proposed method are named Resolution Switchable Networks (RS-Nets). The basic training framework shares network parameters for handling images which differ in resolution, yet keeps separate batch normalization layers. Though it is parameter-efficient in design, it leads to inconsistent accuracy variations at different resolutions, for which we provide a detailed analysis from the aspect of the train-test recognition discrepancy. A multi-resolution ensemble distillation is further designed, where a teacher is learnt on the fly as a weighted ensemble over resolutions. Thanks to the ensemble and knowledge distillation, RS-Nets  enjoy accuracy improvements at a wide range of resolutions compared with individually trained models. Extensive experiments on the ImageNet dataset are provided, and we additionally consider quantization problems. Code and models are available at \url{https://github.com/yikaiw/RS-Nets}.
\end{abstract}

\section{Introduction}
\label{intro}
Convolutional Neural Networks (CNNs) have achieved great success on image recognition tasks \cite{DBLP:conf/cvpr/HeZRS16,DBLP:conf/nips/KrizhevskySH12}, and well-trained recognition models usually need to be deployed on mobile phones, robots or autonomous vehicles \cite{DBLP:conf/iclr/CaiZH19,DBLP:journals/corr/HowardZCKWWAA17}. To fit the resource constraints of devices, extensive research efforts have been devoted to balancing between accuracy and efficiency, by reducing computational complexities of models. Some of these methods adjust the structural configurations of networks, e.g., by adjusting the network depths \cite{DBLP:conf/cvpr/HeZRS16}, widths \cite{DBLP:journals/corr/HowardZCKWWAA17,DBLP:conf/iclr/YuYXYH19} or the convolutional blocks \cite{DBLP:conf/eccv/MaZZS18,DBLP:conf/cvpr/ZhangZLS18}. Besides that, adjusting the image resolution is another widely-used method for the accuracy-efficiency trade-off \cite{DBLP:journals/corr/abs-1905-02244,DBLP:journals/corr/HowardZCKWWAA17,DBLP:journals/corr/abs-1908-03888,DBLP:conf/cvpr/SandlerHZZC18}. If input images are downsized, all feature resolutions at different convolutional layers are reduced subsequently with the same ratio, and the computational cost of a model is nearly proportional to the image resolution ($H\times W$) \cite{DBLP:journals/corr/HowardZCKWWAA17}. However, for a common image recognition model, when the test image resolution differs from the resolution used for training, the accuracy quickly deteriorates \cite{DBLP:journals/corr/abs-1906-06423}. To address this issue, existing works \cite{DBLP:journals/corr/abs-1905-02244,DBLP:conf/cvpr/SandlerHZZC18,DBLP:conf/icml/TanL19} train an individual model for each resolution. As a result, the total number of models to be trained and saved is proportional to the amount of resolutions considered at runtime. Besides the high storage costs, each time adjusting the resolution is accompanied with the additional latency to load another model which is trained with the target resolution.

The ability to switch the image resolution at inference meets a common need for real-life model deployments. By switching resolutions, the running speeds and costs are adjustable to flexibly handle the real-time latency and power requirements for different application scenarios or workloads. Besides, the flexible latency compatibility allows such model to be deployed on a wide range of resource-constrained platforms, which is friendly for application developers. In this paper, we focus on switching input resolutions for an image recognition model, and propose a general and economic method to improve overall accuracies. Models trained with our method are called \textbf{Resolution Switchable Networks (RS-Nets)}. Our contribution is composed of three parts.

\begin{figure}[t]
\centering
\includegraphics[scale=0.5]{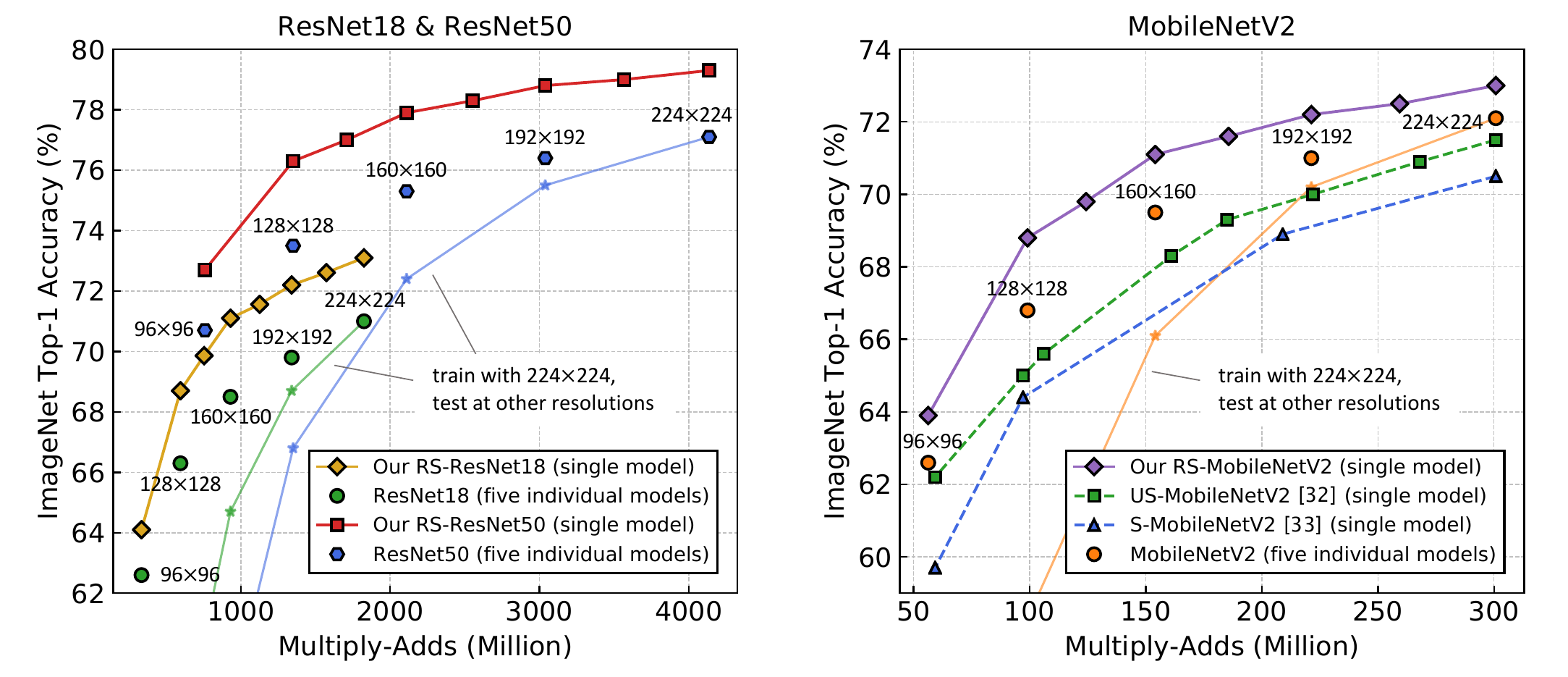}
\vskip-0.8em
\label{mflops_all_nets}
\caption{ImageNet accuracy vs. FLOPs (Multiply-Adds) of our \textbf{single} models and the corresponding \textbf{sets} of individual models. A single RS-Net model is executable at each of the resolutions, and even achieves significantly higher accuracies than individual models. The results of two state-of-the-art switchable networks (switch by varying network widths) S-MobileNetV2 \cite{DBLP:conf/iclr/YuYXYH19} and US-MobileNetV2 \cite{DBLP:journals/corr/abs-1903-05134} are provided for comparison.
Details are in Table \ref{tabs:main_results} and Table \ref{tabs:mflops}.} 
\vskip-1.4em
\end{figure}

First, we propose a parallel training framework where images with different resolutions are trained within a single model. As the resolution difference usually leads to the difference of activation statistics in a network \cite{DBLP:journals/corr/abs-1906-06423}, we adopt shared network parameters but privatized Batch Normalization layers (BNs) \cite{DBLP:conf/icml/IoffeS15} for each resolution. Switching BNs enables the model to flexibly switch image resolutions, without needing to adjust other network parameters.

Second, we associate the multi-resolution interaction effects with a kind of train-test discrepancy (details in Section \ref{sec:discrepancy}). Both our analysis and empirical results reach an interesting conclusion that the parallel training framework tends to enlarge the accuracy gaps over different resolutions. On the one hand, accuracy promotions at high resolutions make a stronger teacher potentially available. On the other hand, the accuracy drop at the lower resolution indicates that the benefits of parallel training itself are limited. Both reasons encourage us to further propose a design of ensemble distillation to improve overall performance.

Third, to the best of our knowledge, we are the first to propose that a data-driven ensemble distillation can be learnt on the fly for image recognition, based on the same image instances with different resolutions. Regarding the supervised image recognition, the structure of our design is also different from existing ensemble or knowledge distillation works, as they focus on the knowledge transfer among different models, e.g., by stacking multiple models \cite{DBLP:conf/cvpr/ZhangXHL18}, multiple branches \cite{DBLP:conf/nips/LanZG18}, pre-training a teacher model, or splitting the model into sub-models \cite{DBLP:journals/corr/abs-1903-05134}, while our model is single and shared, with little extra parameters.

Extensive experiments on the ImageNet dataset validate that RS-Nets are executable given different image resolutions at runtime, and achieve significant accuracy improvements at a wide range of resolutions compared with individually trained models. Illustrative results are provided in Fig. \ref{mflops_all_nets}, which also verify that our proposed method can be generally applied to modern recognition backbones.

\section{Related Work}
\label{rw}
\subsubsection{Image Recognition.}  Image recognition acts as a benchmark to evaluate models and is a core task for computer vision. Advances on the large-scale image recognition datasets, like ImageNet \cite{DBLP:conf/cvpr/DengDSLL009}, can translate to improved results on a number of other applications \cite{DBLP:journals/ijcv/GordoARL17,DBLP:conf/cvpr/KornblithSL19,DBLP:conf/cvpr/OquabBLS14}. 
In order to enhance the model generalization for image recognition, data augmentation strategies, e.g., random-size crop, are adopted during training \cite{DBLP:conf/cvpr/HeZRS16,DBLP:journals/corr/abs-1812-01187,DBLP:conf/cvpr/SzegedyLJSRAEVR15}. Besides, common models are usually trained and tested with fixed-resolution inputs. \cite{DBLP:journals/corr/abs-1906-06423} shows that for a model trained with the default $224\times224$ resolution and tested at lower resolutions, the accuracy quickly deteriorates (e.g., drops 11.6\% at test resolution $128\times128$ on ResNet50). 

\subsubsection{Accuracy-Efficiency Trade-Off.} 
\vskip-1em There have been many attempts to balance accuracy and efficiency by model scaling. Some of them adjust the structural configurations of networks. For example, 
ResNets \cite{DBLP:conf/cvpr/HeZRS16} provide several choices of network depths from shallower to deep. MobileNets \cite{DBLP:journals/corr/HowardZCKWWAA17,DBLP:conf/cvpr/SandlerHZZC18} and ShuffleNets \cite{DBLP:conf/cvpr/ZhangZLS18} can reduce network widths by using smaller width multipliers. While some other works \cite{DBLP:journals/corr/abs-1905-02244,DBLP:journals/corr/HowardZCKWWAA17,DBLP:journals/corr/abs-1908-03888,DBLP:conf/cvpr/SandlerHZZC18} reduce the computational complexity by decreasing image resolutions at input, which is also our focus. Modifying the resolution usually does not make changes to the number of network parameters, but significantly affects the computational complexity \cite{DBLP:journals/corr/HowardZCKWWAA17}.

\subsubsection{Knowledge Distillation.} 
\vskip-1em 
A student network can be improved by imitating feature representations or soft targets of a larger teacher network \cite{DBLP:journals/corr/HintonVD15,DBLP:conf/eccv/LiH16,DBLP:conf/cvpr/ZhangXHL18}. The teacher is usually pre-trained beforehand and fixed, and the knowledge is transferred in one direction \cite{DBLP:journals/corr/RomeroBKCGB14}. Yet \cite{DBLP:conf/cvpr/ZhangXHL18} introduces a two-way transfer between two peer models. \cite{DBLP:conf/cvpr/SunYZZ19} performs mutual learning within one single network assisted by intermediate classifiers. \cite{DBLP:conf/nips/LanZG18} learns a native ensemble design based on multiple models for distillation. \cite{DBLP:journals/corr/abs-1903-05134} conducts the knowledge distillation between the whole model and each split smaller model. Regarding the supervised image recognition, existing distillation works rely on different models, usually needing another teacher network with higher-capacity than low-capacity students. While our design is applied in a shared model, which is data-driven, collecting complementary knowledge from the same image instances with different resolutions.

\section{Proposed Method}
\begin{figure}[t]
\centering
\hskip-0.3em
\resizebox{\linewidth}{!}{
\includegraphics[scale=0.42]{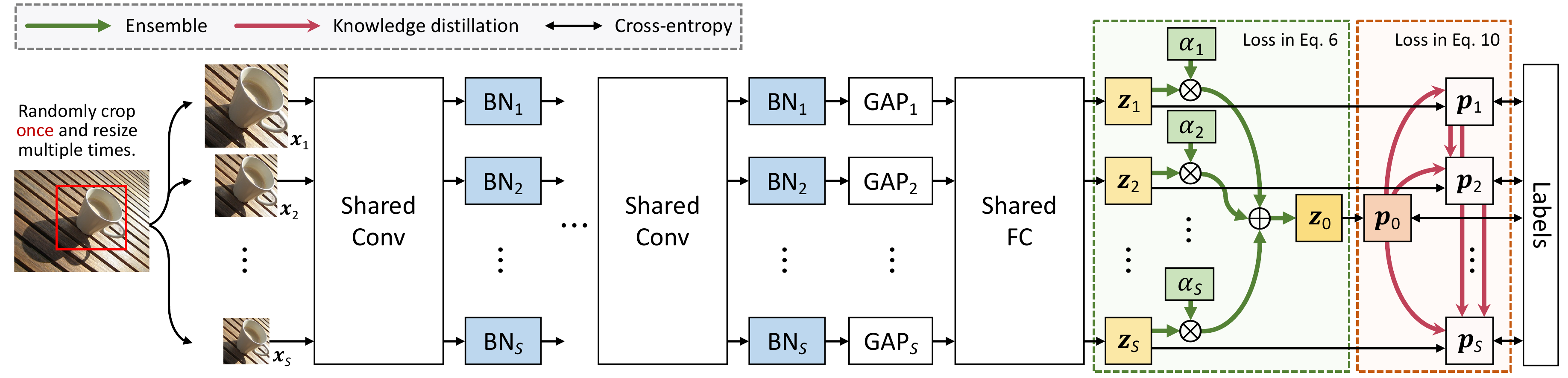}}
\caption{Overall framework of training a RS-Net. Images with different resolutions are trained in parallel with shared Conv/FC layers and private BNs. The ensemble logit ($\bm{z}_0$) is learnt on the fly as a weighted mean of logits ($\bm{z}_1,\bm{z}_2,\cdots,\bm{z}_S$), shown as green arrows. Knowledge distillations are shown as red arrows. For inference, one of the $S$ forward paths is selected (according to the image resolution), with its corresponding BNs, for obtaining its corresponding prediction $\bm{p}_s, s\in\{1,2,\cdots,S\}$. \emph{The ensemble and knowledge distillation are not needed during inference.}}
\label{framework}
\vskip-0.7em
\end{figure}

\vskip-0.1em
A schematic overview of our proposed method is shown in Fig. \ref{framework}. In this section, we detail the insights and formulations of parallel training, interaction effects and ensemble distillation based on the multi-resolution setting.

\subsection{Multi-Resolution Parallel Training}
\label{sec:parallel_training}

To make the description self-contained, we begin with the basic training of a CNN model. Given training samples, we crop and resize each sample to a fixed-resolution image $\bm{x}^i$. We denote network inputs as $\{(\bm{x}^i,y^i)|i\in\{1,2,\cdots,N\}\}$, where $y^i$ is the ground truth which belongs to one of the $C$ classes, and $N$ is the amount of samples. Given network configurations with parameters $\bm{\theta}$, the predicted probability of the class $c$ is denoted as $p(c|\bm{x}^i, \bm{\theta})$. The model is optimized with a cross-entropy loss defined as:
\vskip-0.3em
\begin{equation}
\label{ce}
\mathcal{H}(\bm{x},\bm{y})=-\frac{1}{N}\sum_{i=1}^N\sum_{c=1}^C\delta(c,y^i)\log\big(p(c|\bm{x}^i, \bm{\theta})\big),
\end{equation}
where $\delta(c,y^i)$ equals to $1$ when $c=y^i$, otherwise $0$.

In this part, we propose \textbf{multi-resolution parallel training}, or called \textbf{parallel training} for brevity, to train a single model which can switch image resolutions at runtime. During training, each image sample is randomly cropped and resized to several duplicate images with different resolutions. Suppose that there are $S$ resolutions in total, the inputs can be written as $\{(\bm{x}_1^i, \bm{x}_2^i, \cdots, \bm{x}_S^i, y^i)|i\in\{1,2,\cdots,N\}\}$. Recent CNNs for image recognition follow similar structures that all stack Convolutional (Conv) layers, a Global Average Pooling (GAP) layer and a Fully-Connected (FC) layer. In CNNs, if input images have different resolutions, the corresponding feature maps in all Conv layers will also vary in resolution. Thanks to GAP, features are transformed to a unified spatial dimension ($1\times1$) with equal amount of channels, making it possible to be followed by a same FC layer. During our parallel training, we share parameters of Conv layers and the FC layer, and therefore the training for multiple resolutions can be realized in a single network. The loss function for parallel training is calculated as a summation of the cross-entropy losses:
\vskip-2mm
\begin{equation}\label{loss_cls}
\mathcal{L}_{cls}=\sum_{s=1}^S \mathcal{H}(\bm{x}_s,\bm{y}).
\end{equation}

Specializing Batch Normalization layers (BNs) \cite{DBLP:conf/icml/IoffeS15} is proved to be effective for efficient model adaption \cite{DBLP:conf/cvpr/ChangYSKH19,DBLP:conf/iclr/MudrakartaSZH19,DBLP:conf/iclr/YuYXYH19,DBLP:journals/corr/abs-1909-00182}. In image recognition tasks, resizing image results in different activation statistics in a network \cite{DBLP:journals/corr/abs-1906-06423}, including means and variances used in BNs. Thus during parallel training, we privatize BNs for each resolution. Results in the left panel of Fig. \ref{share} verify the necessity of privatizing BNs. For the $s^{\text{th}}$ resolution, each corresponding BN layer normalizes the channel-wise feature as follows:
\begin{equation}\label{sbn}
\bm{y}'_s=\bm{\gamma}_s\frac{\bm{y}_s-\bm{\mu}_s}{\sqrt{\bm{\sigma}_s^2+\epsilon}}+\bm{\beta}_s, s\in\{1,2,\cdots,S\},
\end{equation}

where $\bm{\mu}_s$ and $\bm{\sigma}_s^2$ are running mean and variance; $\bm{\gamma}_s$ and $\bm{\beta}_s$ are learnable scale and bias. Switching these parameters enables the model to switch resolutions.

\subsection{Multi-Resolution Interaction Effects}
\label{sec:discrepancy}

\begin{figure}[t]
\centering
\resizebox{115mm}{!}{
\includegraphics[scale=0.13]{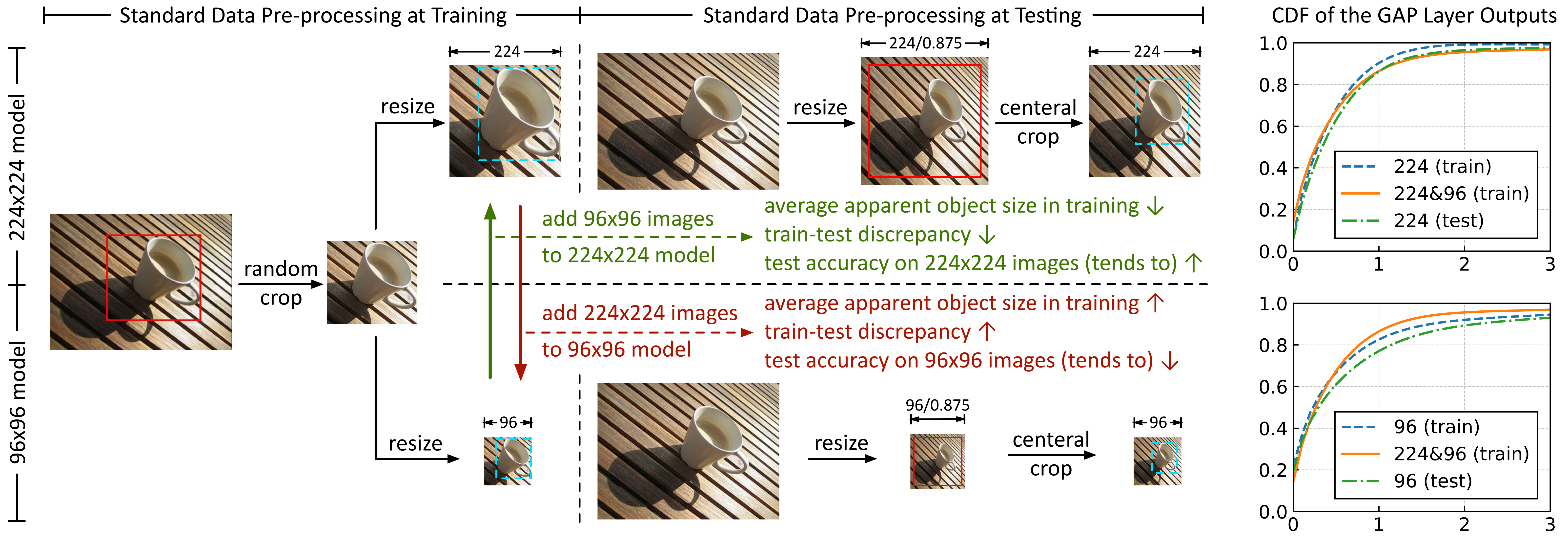}}
\vskip-0.5em
\caption{\textbf{Left:} An illustration of interaction effects for the parallel training with two resolutions. Each red box indicates the region to be cropped, and the size of each blue dotted box is the apparent object (in this sample is a cup) size. For this example, in either one of the models, the apparent object size at testing is smaller than at training. \cite{DBLP:journals/corr/abs-1906-06423} reveals that this relation still holds when averaging all data, which is called the train-test discrepancy. The data pre-processing for training or testing follows the standard image recognition method, which will be described in Section \ref{details}. \textbf{Right:} CDF curves for comparing the value distributions of feature activations. All curves are plotted on the validation dataset of ImageNet, but are based on different data pre-processing methods as annotated by (train) or (test).}
\label{pic_crop}
\vskip-1.4em
\end{figure}

In this section, restricted to large-scale image datasets with fine-resolution images, we analyze the interaction effects of different resolutions under the parallel training framework. We start by posing a question: compared with individually trained models, how does parallel training affect test accuracies at different resolutions? As multi-resolution can be seen as a kind of data augmentation, we analyze from two aspects as follows.

The first aspect is straightforward. The model meets a wide range of image resolutions, which improves the generalization and reduces over-fitting. Thus if the setting of resolutions is suitable (e.g., not too diverse), the parallel training tends to bring overall accuracy gains at testing, especially for a high-capacity network such as ResNet50.

The second aspect is based on the specialty of large-scale image recognition tasks, where objects of interest randomly occupy different portions of image areas and thus the random-size crop augmentation is used during training. \cite{DBLP:journals/corr/abs-1906-06423} reveals that for common recognition works, as the random-size crop is used for training but not for testing, there exists a train-test \textbf{discrepancy} that the average ``apparent object size" at testing is smaller than that at training. Besides, \cite{DBLP:journals/corr/abs-1906-06423} achieves accuracy improvements by alleviating such discrepancy, but is accompanied with the costs of test-time resolution augmentations and finetuning. Note that we do not aim to modify the data pre-processing method to alleviate the discrepancy. Instead, we are inspired to use the concept of the discrepancy to analyze multi-resolution interaction effects. We take the parallel training with two resolutions $224\times224$ and $96\times96$ for example. According to the left panel of Fig. \ref{pic_crop} (the analysis in colors), compared with the model using only $224\times224$ images, the parallel training can be seen as augmenting $96\times96$ images, which reduces the average apparent object size at training and thus alleviates the discrepancy. On the contrary, compared with the individual model using $96\times96$ images, augmenting $224\times224$ images increases the discrepancy. Thus in this aspect, this parallel training tends to increase the test accuracy at $224\times224$ (actually +1.6\% for ResNet18, as shown in Fig. \ref{parallel}), while tends to reduce the test accuracy at $96\times96$ (-0.8\%). The right panel of Fig. \ref{pic_crop} plots the Cumulative Distribution Function (CDF)\footnote{The CDF of a random variable $X$ is defined as $F_X(x)=P(X\le x)$, for all $x\in \mathbb{R}$.} of output components of the Global Average Pooling (GAP) layer for a well-trained ResNet18 (as a ReLU layer is before the GAP, all components are nonnegative). We plot CDF to compare the value distributions of feature activations when using training or testing data pre-processing method. The parallel training seems to narrow the train-test gap for the $224\times224$ model, but widen the gap for the $96\times96$ model.

We take ResNet18 as an example and summarize the two aforementioned aspects. For the parallel training with two resolutions, the test accuracy at the high resolution increases compared with its individual model, as the two aspects reach an agreement to a large degree. As for the lower resolution, we find that the accuracy slightly increases if the two resolutions are close, otherwise decreases. Similarly, when multiple resolutions are used for parallel training, test accuracies increase at high resolutions but may decrease at lower resolutions, compared with individual models. Results in Table \ref{tabs:main_results} show that for the parallel training with five resolutions, accuracies only decrease at $96\times96$ but increase at the other four resolutions. Detailed results in Fig. \ref{parallel} also verify our analysis.

For image recognition, although testing at a high resolution already tends to achieve a good accuracy, using the parallel training makes it even better. This finding opens up a possibility that a stronger teacher may be available in this framework, and seeking a design based on such teacher could be highly effective.

\subsection{Multi-Resolution Ensemble Distillation}
\label{mred}

In this section, we propose a new design of ensemble distillation. Regarding the supervised image recognition, unlike conventional  distillation works that rely on transferring knowledge among different models, ours is data-driven and can be applied in a shared model. Specifically, our design is learnt on the fly and the distillation is based on the same image instances with different resolutions.

\begin{table}[t]

\begin{center}
\caption{Proportions (\%) of validation samples that are correctly classified ($\surd$) at a resolution but are wrongly classified ($\times$) at another, based on ResNet18. All models are well-trained. Lower numbers correspond to better performance, and numbers which are larger than the base are colored red, otherwise green. Benefited from MRED, all the proportions in the last column decrease.}
\vskip-0.3em
\hskip-0.1em
\resizebox{120mm}{!}{
\setlength{\tabcolsep}{1.5mm}{
\begin{tabular}{p{0.8cm}<{\centering}|p{0.6cm}<{\centering}p{0.6cm}<{\centering}p{0.6cm}<{\centering}p{0.6cm}<{\centering}p{0.6cm}<{\centering}|p{0.9cm}<{\centering}p{0.9cm}<{\centering}p{0.9cm}<{\centering}p{0.8cm}<{\centering}p{0.8cm}<{\centering}|p{0.9cm}<{\centering}p{0.9cm}<{\centering}p{0.9cm}<{\centering}p{0.8cm}<{\centering}p{0.8cm}<{\centering}}
\toprule[1.2pt]
\multirow{2}*{\diagbox [width=3em] {$\times$}{$\surd$} } & \multicolumn{5}{c|}{Individual Training (base)} &  \multicolumn{5}{c|}{Parallel Training} &  \multicolumn{5}{c}{Parallel Training + MRED}\\

&$224$ & $192$ & $160$ & $128$ & $96$&$224$ & $192$ & $160$ & $128$ & $96$&$224$ & $192$ & $160$ & $128$ & $96$\\
\midrule[0.7pt]
$224$&-&5.7&5.5&5.3&4.7&-&\textcolor[RGB]{53, 134, 20}{3.0 $\bm{\downarrow}$}&\textcolor[RGB]{53, 134, 20}{3.4 $\bm{\downarrow}$}&\textcolor[RGB]{53, 134, 20}{3.8 $\bm{\downarrow}$}&\textcolor[RGB]{53, 134, 20}{3.9 $\bm{\downarrow}$}&-&\textcolor[RGB]{53, 134, 20}{2.5 $\bm{\downarrow}$}&\textcolor[RGB]{53, 134, 20}{2.9 $\bm{\downarrow}$}&\textcolor[RGB]{53, 134, 20}{3.4 $\bm{\downarrow}$}&\textcolor[RGB]{53, 134, 20}{3.6 $\bm{\downarrow}$}\\
$192$&6.9&-&6.0&5.5&5.2&\textcolor[RGB]{53, 134, 20}{4.1 $\bm{\downarrow}$}&-&\textcolor[RGB]{53, 134, 20}{3.3 $\bm{\downarrow}$}&\textcolor[RGB]{53, 134, 20}{3.7 $\bm{\downarrow}$}&\textcolor[RGB]{53, 134, 20}{3.8 $\bm{\downarrow}$}&\textcolor[RGB]{53, 134, 20}{3.4 $\bm{\downarrow}$}&-&\textcolor[RGB]{53, 134, 20}{2.8 $\bm{\downarrow}$}&\textcolor[RGB]{53, 134, 20}{3.3 $\bm{\downarrow}$}&\textcolor[RGB]{53, 134, 20}{3.4 $\bm{\downarrow}$}\\
$160$&8.1&7.2&-&5.9&5.4&\textcolor[RGB]{53, 134, 20}{6.0 $\bm{\downarrow}$}&\textcolor[RGB]{53, 134, 20}{4.7 $\bm{\downarrow}$}&-&\textcolor[RGB]{53, 134, 20}{3.5 $\bm{\downarrow}$}&\textcolor[RGB]{53, 134, 20}{3.6 $\bm{\downarrow}$}&\textcolor[RGB]{53, 134, 20}{5.1 $\bm{\downarrow}$}&\textcolor[RGB]{53, 134, 20}{3.9 $\bm{\downarrow}$}&-&\textcolor[RGB]{53, 134, 20}{2.9 $\bm{\downarrow}$}&\textcolor[RGB]{53, 134, 20}{3.2 $\bm{\downarrow}$}\\
$128$&9.8&8.9&8.1&-&5.8&\textcolor[RGB]{53, 134, 20}{9.3 $\bm{\downarrow}$}&\textcolor[RGB]{53, 134, 20}{8.2 $\bm{\downarrow}$}&\textcolor[RGB]{53, 134, 20}{6.6 $\bm{\downarrow}$}&-&\textcolor[RGB]{53, 134, 20}{3.5 $\bm{\downarrow}$}&\textcolor[RGB]{53, 134, 20}{7.3 $\bm{\downarrow}$}&\textcolor[RGB]{53, 134, 20}{6.4 $\bm{\downarrow}$}&\textcolor[RGB]{53, 134, 20}{5.0 $\bm{\downarrow}$}&-&\textcolor[RGB]{53, 134, 20}{3.1 $\bm{\downarrow}$}\\
$96$&13.2&12.5&11.4&9.7&-&\textcolor{red}{15.2 $\bm{\uparrow}$}&\textcolor{red}{14.2 $\bm{\uparrow}$}&\textcolor{red}{12.7 $\bm{\uparrow}$}&\textcolor{red}{9.8 $\bm{\uparrow}$}&-&\textcolor[RGB]{53, 134, 20}{12.2 $\bm{\downarrow}$}&\textcolor[RGB]{53, 134, 20}{11.4 $\bm{\downarrow}$}&\textcolor[RGB]{53, 134, 20}{10.2 $\bm{\downarrow}$}&\textcolor[RGB]{53, 134, 20}{7.9 $\bm{\downarrow}$}&-\\
\bottomrule[1.2pt]
\end{tabular}

\label{tab_proportion}}
}
\end{center}

\vskip-2.3em
\end{table}

As is commonly known, for image recognition tasks, models given a high resolution image are easy to capture fine-grained patterns, and thus achieve good performance \cite{DBLP:conf/cvpr/SzegedyVISW16,DBLP:conf/icml/TanL19}. However, according to the sample statistics in the middle column of Table \ref{tab_proportion}, we find that there always exists a proportion of samples which are correctly classified at a low resolution but wrongly classified at another higher resolution. Such results indicate that model predictions at different image resolutions are complementary, and not always the higher resolution is better for each image sample. Therefore, we propose to learn a teacher on the fly as an ensemble of the predictions w.r.t. all resolutions, and conduct knowledge distillation to improve the overall performance. Our design is called \textbf{Multi-Resolution Ensemble Distillation (MRED)}.

During the training process of image recognition, for each input image $\bm{x}^i$, the probability of the class $c$ is calculated using a softmax function:\vskip-0.2em
\begin{equation}\label{p}
p(c|\bm{x}^i, \bm{\theta})=p(c|\bm{z}^i)=\frac{\exp(z_c^i)}{\sum_{j=1}^{C}\exp(z_j^i)}, c\in\{1,2,\cdots,C\},
\end{equation}
where $\bm{z}^i$ is the logit, the unnormalized log probability outputted by the network, and probabilities over all classes can be denoted as the model prediction $\bm{p}$.

In the parallel training framework, each image is randomly cropped and resized to $S$ images with different resolutions. To better benefit from MRED, these $S$ images need to be resized from a same random crop, as illustrated in the left-most part of Fig. \ref{framework}. The necessity will be verified in Section \ref{veri_kd}.

As each image sample is resized to $S$ resolutions, there are $S$ corresponding logits $\bm{z}_1,\bm{z}_2,\cdots,\bm{z}_S$. We learn a group of importance scores $\bm{\alpha}\!=[\alpha_1\,\alpha_2\,\cdots\,\alpha_S]$, satisfying $\bm{\alpha}\!\ge\!0$, $\sum_{s=1}^S\alpha_s\!=\!1$, which can be easily implemented with a softmax function. We then calculate an ensemble logit $\bm{z}_0$ as the weighted summation of the $S$ logits:
\begin{equation}\label{ens_logits}
\hskip-0.8em
\bm{z}_0=\small{\sum_{s=1}^S}\alpha_s \bm{z}_s.
\end{equation}

To optimize $\bm{\alpha}$, we temporally froze the gradients of the logits $\bm{z}_1,\bm{z}_2,\cdots,\bm{z}_S$. Based on the ensemble logit $\bm{z}_0$, the corresponding prediction $\bm{p}_0$, called ensemble prediction,  can be calculated via Eq. \ref{p}. Then $\bm{\alpha}$ is optimized using a cross-entropy loss between $\bm{p}_0$ and the ground truth, which we call the ensemble loss $\mathcal{L}_{ens}$:
\begin{equation}\label{loss_ens}
\mathcal{L}_{ens}=-\frac{1}{N}\sum_{i=1}^N\sum_{c=1}^C\delta(c,y^i)\log\big(p(c|\bm{z}_0^i)\big).
\end{equation}

In knowledge distillation works, to quantify the alignment between a teacher prediction $\bm{p}_t$ and a student prediction $\bm{p}_s$, Kullback Leibler (KL) divergence is usually used:
\begin{equation}\label{kl}
\mathcal{D}_{kl}\big(\bm{p}_t\|\bm{p}_s\big)=\frac{1}{N}\sum_{i=1}^N\sum_{c=1}^C p(c|\bm{z}_t^i)\log\frac{p(c|\bm{z}_t^i)}{p(c|\bm{z}_s^i)}.
\end{equation}

We force predications at different resolutions to mimic the learnt ensemble prediction $\bm{p}_0$, and thus the distillation loss $\mathcal{L}_{dis}$ could be obtained as:
\begin{equation}\label{loss_kd_v1}
\mathcal{L}_{dis}=\sum_{s=1}^S\mathcal{D}_{kl}\big(\bm{p}_0\|\bm{p}_s\big).
\end{equation}

Finally, the overall loss function is a summation of the classification loss, the ensemble loss and the distillation loss, without needing to tune any extra weighted parameters:
\begin{equation}\label{loss}
\mathcal{L}=\mathcal{L}_{cls}+\mathcal{L}_{ens}+\mathcal{L}_{dis},
\end{equation}
where in practical, optimizing $\mathcal{L}_{ens}$ only updates $\bm{\alpha}$, with all network weights temporally frozen; optimizing $\mathcal{L}_{cls}$ and $\mathcal{L}_{dis}$ updates network weights. 

We denote the method with Eq. \ref{loss_kd_v1} as our vanilla-version MRED. Under the parallel training framework, as the accuracy at a high resolution is usually better than at a lower resolution, accuracies can be further improved by offering dense guidance from predications at high resolutions toward predictions at lower resolutions. Thus the distillation loss can be extended to be a generalized one:\vskip-0.2em
\begin{equation}\label{loss_kd_v2}
\mathcal{L}_{dis}=\frac{2}{S+1}\sum_{t=0}^{S-1}\sum_{s=t+1}^S\mathcal{D}_{kl}\big(\bm{p}_t\|\bm{p}_s\big),
\end{equation}
where the index $t$ starts from $0$ referring to the ensemble term; as the summation results in $S(S+1)/2$ components in total, we multiply $\mathcal{L}_{dis}$ by a constant ratio $2/(S+1)$ to keep its range the same as $\mathcal{L}_{cls}$. We denote the method with Eq. \ref{loss_kd_v2} as our full-version MRED, which is used in our experiments by default.

The proposed design involves negligible extra parameters (only $S$ scalars), without needing extra models. Models trained with the parallel training framework and MRED are named Resolution Switchable Networks (RS-Nets). An overall framework of training a RS-Net is illustrated in Fig. \ref{framework}. \textbf{During inference}, the network only performs one forward calculation at a given resolution, without ensemble or distillation, and thus both the computational complexity and the amount of parameters equal to a conventional image recognition model.

\section{Experiments}
We perform experiments on ImageNet (ILSVRC12) \cite{DBLP:conf/cvpr/DengDSLL009,DBLP:journals/ijcv/RussakovskyDSKS15}, a widely-used image recognition dataset containing about 1.2 million training images and 50 thousand validation images, where each image is annotated as one of 1000 categories. Experiments are conducted with prevailing CNN architectures  including a lightweight model MobileNetV2 \cite{DBLP:conf/cvpr/SandlerHZZC18} and ResNets \cite{DBLP:conf/cvpr/HeZRS16}, where a basic-block model ResNet18 and a bottleneck-block model ResNet50 are both considered. Besides, we also evaluate our method in handling network quantization problems, where we consider different kinds of bit-widths.

\subsection{Implementation Details}
\label{details}
Our basic experiments are implemented with PyTorch$\,$\cite{DBLP:conf/nips/PaszkeGMLBCKLGA19}.$\,$For quantization tasks, we apply our method to LQ-Nets \cite{DBLP:conf/eccv/ZhangYYH18} which show state-of-the-art performance in training CNNs with low-precision weights or both weights and activations. 

We set $\mathbb{S}=\{224\times224,192\times192,160\times160,128\times128,96\times96\}$, as commonly adopted in a number of existing works \cite{DBLP:journals/corr/HowardZCKWWAA17,DBLP:journals/corr/abs-1908-03888,DBLP:conf/cvpr/SandlerHZZC18}. During training, we pre-process the data for augmentation with an area ratio ($\text{cropped area}/\text{original area}$) uniformly sampled in $[0.08, 1.0]$, an aspect ratio $[3/4, 4/3]$ and a horizontal flipping. We resize images with the bilinear interpolation. Note that both $[0.08, 1.0]$ and $[3/4, 4/3]$ follow the standard data augmentation strategies for ImageNet \cite{DBLP:conf/cvpr/HuangLMW17,DBLP:conf/cvpr/SzegedyLJSRAEVR15,DBLP:journals/corr/abs-1906-06423}, e.g., \emph{RandomResizedCrop} in PyTorch uses such setting as default. During validation, we first resize images with the bilinear interpolation to every resolution in $\mathbb{S}$ divided by 0.875 \cite{DBLP:journals/corr/abs-1908-08986,DBLP:journals/corr/abs-1908-03888}, and then feed central regions to models.

Networks are trained from scratch with random initializations. We set the batch size to 256, and use a SGD optimizer with a momentum 0.9. For standard ResNets, we train 120 epochs and the learning rate is annealed from 0.1 to 0 with a cosine scheduling \cite{DBLP:journals/corr/abs-1812-01187}. For MobileNetV2, we train 150 epochs and the learning rate is annealed from 0.05 to 0 with a cosine scheduling. For quantized ResNets, we follow the settings in LQ-Nets \cite{DBLP:conf/eccv/ZhangYYH18}, which train 120 epochs and the learning rate is initialized to 0.1 and divided by 10 at 30, 60, 85, 95, 105 epochs. The weight decay rate is set to 1e-4 for all ResNets and 4e-5 for MobileNetV2.

\subsection{Results}
As mentioned in Section \ref{intro}, common works with multi-resolution settings train and deploy multiple individual models separately for different resolutions. We denote these individual models as \textbf{I-Nets}, which are set as baselines. We use I-\{resolution\} to represent each individual model, e.g., I-$224$.

\newcommand{\tcc}[1]{\multicolumn{1}{c|}{#1}}
\newcommand{\tc}[1]{\multicolumn{1}{c}{#1}}

\begin{table}[t]
\centering
\caption{Basic results comparison on ImageNet. We report top-1/top-5 accuracies (\%), top-1 accuracy gains (\%) over individual models (I-Nets), Multiply-Adds (MAdds) and total parameters (params). All experiments use the same data pre-processing methods. Our baseline results are slightly higher than the original papers \cite{DBLP:conf/cvpr/HeZRS16,DBLP:conf/cvpr/SandlerHZZC18}.}
\vskip-0.3em
\resizebox{108mm}{!}{
\begin{tabular}{p{1.5cm}<{\centering}|p{1.7cm}<{\centering}|p{1.2cm}<{\centering}|p{1.8cm}<{\centering}p{0.cm}lp{0.1cm}lp{0.1cm}lp{0.07cm}} 
\toprule[1.2pt]
Network & Resolution & MAdds & \tc{I-Nets (base)} && \tc{I-$224$} && \tc{Our Parallel} && \tc{Our RS-Net} &\\
\midrule[0.7pt]
\multirow{6}{*}{ResNet18}
& $224\times224$ & 1.82G & 71.0 / 90.0 && 71.0 / 90.0 && 73.0 / 90.9 \textsubscript{(+2.0)} && \textbf{73.1 / 91.0 \textsubscript{(+2.1)}} &\\
& $192\times192$ & 1.34G & 69.8 / 89.4 && 68.7 / 88.5 \textsubscript{(-1.1)} && 71.7 / 90.3 \textsubscript{(+1.9)} && \textbf{72.2 / 90.6 \textsubscript{(+2.4)}} &\\
& $160\times160$ & 931M & 68.5 / 88.2 && 64.7 / 85.9 \textsubscript{(-5.2)} && 70.4 / 89.6 \textsubscript{(+1.9)} && \textbf{71.1 / 90.1 \textsubscript{(+2.6)}} &\\
& $128\times128$ & 596M & 66.3 / 86.8 && 56.8 / 80.0 \textsubscript{(-9.5)} && 67.5 / 87.8 \textsubscript{(+1.2)} && \textbf{68.7 / 88.5 \textsubscript{(+2.4)}} &\\
& $96\times96$ & 335M & 62.6 / 84.1 && 42.5 / 67.9 \textsubscript{(-20.1)} && 61.5 / 83.5 \textsubscript{(-1.1)} && \textbf{64.1 / 85.3 \textsubscript{(+1.5)}} & \\
\cmidrule(r){2-11}
& \multicolumn{2}{c|}{Total Params} & \tc{55.74M} && \tc{11.15M} && \tc{11.18M}  && \tc{11.18M} &\\
\midrule[0.7pt]

\multirow{6}{*}{ResNet50} 
& $224\times224$ & 4.14G & 77.1 / 93.4 && 77.1 / 93.4 && 78.9 / 94.4 \textsubscript{(+1.8)} && \textbf{79.3 / 94.6 \textsubscript{(+2.2)}} &\\
& $192\times192$ & 3.04G & 76.4 / 93.2 && 75.5 / 92.5  \textsubscript{(-0.9)} && 78.1 / 94.0 \textsubscript{(+1.7)} && \textbf{78.8 / 94.4 \textsubscript{(+2.4)}} &\\
& $160\times160$ & 2.11G & 75.3 / 92.4 && 72.4 / 90.7  \textsubscript{(-2.9)} && 76.9 / 93.1 \textsubscript{(+1.6)} && \textbf{77.9 / 93.9 \textsubscript{(+2.6)}} &\\
& $128\times128$ & 1.35G & 73.5 / 91.4 &&  66.8 / 87.0 \textsubscript{(-6.7)} && 74.9 / 92.1 \textsubscript{(+1.4)} && \textbf{76.3 / 93.0 \textsubscript{(+2.8)}} &\\
& $96\times96$ & 760M & 70.7 / 89.8 && 54.9 / 78.2 \textsubscript{(-15.8)} && 70.2 / 89.4 \textsubscript{(-0.5)} && \textbf{72.7 / 91.0 \textsubscript{(+2.0)}} & \\
\cmidrule(r){2-11}
& \multicolumn{2}{c|}{Total Params} & \tc{121.87M} && \tc{24.37M} && \tc{24.58M}  && \tc{24.58M} &\\
\midrule[0.7pt]

\multirow{6}{*}{M-NetV2} 
& $224\times224$ & 301M & 72.1 / 90.5 && 72.1 / 90.5 && 72.8 / 90.9 \textsubscript{(+0.7)} && \textbf{73.0 / 90.8 \textsubscript{(+0.9)}} &\\
& $192\times192$ & 221M & 71.0 / 89.8 && 70.2 / 89.1 \textsubscript{(-0.9)} && 71.7 / 90.2 \textsubscript{(+0.7)} && \textbf{72.2 / 90.5 \textsubscript{(+1.2)}} &\\
& $160\times160$ & 154M & 69.5 / 88.9 && 66.1 / 86.3 \textsubscript{(-3.2)} && 70.1 / 89.2 \textsubscript{(+0.6)} && \textbf{71.1 / 90.2 \textsubscript{(+1.6)}} &\\
& $128\times128$ & 99M & 66.8 / 87.0 && 58.3 / 81.2 \textsubscript{(-8.5)} && 67.3 / 87.2 \textsubscript{(+0.5)} && \textbf{68.8 / 88.2 \textsubscript{(+2.0)}} &\\
& $96\times96$ & 56M & 62.6 / 84.0 && 43.9 / 69.1 \textsubscript{(-18.7)} && 61.4 / 83.3 \textsubscript{(-1.2)} && \textbf{63.9 / 84.9 \textsubscript{(+1.3)}} & \\
\cmidrule(r){2-11}
& \multicolumn{2}{c|}{Total Params} & \tc{16.71M} && \tc{3.34M}  && \tc{3.47M} && \tc{3.47M} &\\

\bottomrule[1.2pt]
\end{tabular}}
\label{tabs:main_results}
\vskip -1em
\end{table}

\subsubsection{Basic Results.} 
\vskip-0.3em
In Table \ref{tabs:main_results}, we report results on ResNet18, ResNet50 and MobileNetV2 (M-NetV2 for short). Besides I-Nets, we also report accuracies at five resolutions using the individual model which is trained with the largest resolution (I-$224$). For our proposed method, we provide separate results of the parallel training (parallel) and the overall design (RS-Net). As mentioned in Section \ref{intro}, I-Nets need several times of parameter amount and high latencies for switching across models. We also cannot rely on an individual model to switch the image resolutions, as accuracies of I-$224$ are much lower than I-Nets at other resolutions (e.g., 15\%$\sim$20\% accuracy drop at the resolution $96\times96$). Similarly, each of the other individual models also suffers from serious accuracy drops, as can be seen in the right panel of Fig. \ref{share}. Our parallel training brings accuracy improvements at the four larger resolutions, while accuracies at $96\times96$ decrease, and the reason is previously analyzed in Section \ref{sec:discrepancy}. Compared with I-Nets, our RS-Net achieves large improvements at all resolutions with only 1$/$5 parameters. For example, the RS-Net with ResNet50 obtains about 2.4\% absolute top-1 accuracy gains on average across five resolutions. Note that the number of FLOPs (Multiply-Adds) is nearly proportional to the image resolution \cite{DBLP:journals/corr/HowardZCKWWAA17}. Regarding ResNet18 and ResNet50, accuracies at $160\times160$ of our RS-Nets even surpass the accuracies of I-Nets at $224\times224$, significantly reducing about 49\% FLOPs at runtime. Similarly, for MobileNetV2, the accuracy at $192\times192$ of RS-Net surpasses the accuracy of I-Nets at $224\times224$, reducing about 26\% FLOPs.

\newcommand{\tabincell}[2]{\begin{tabular}{@{}#1@{}}#2\end{tabular}}

\begin{table}[t]
\centering

\caption{Results comparison for quantization tasks. We report top-1/top-5 accuracies (\%) and top-1 accuracy gains (\%) over individual models (I-Nets). All experiments are performed under the same training settings following LQ-Nets.}
\vskip-0.1em
\resizebox{108mm}{!}{
\begin{tabular}{p{1.7cm}<{\centering}|p{1.7cm}<{\centering}|p{2cm}<{\centering}lp{0.1cm}<{\centering}|p{2cm}<{\centering}lp{0.1cm}<{\centering}} 
\toprule[1.2pt]
\multirow{2}*{Network} & \multirow{2}*{Resolution} & \multicolumn{3}{c|}{Bit-width (W/A): 2 / 32}& \multicolumn{3}{c}{Bit-width (W/A): 2 / 2} \\
&&\tc{I-Nets (base)} & \tc{Our RS-Net} && \tc{I-Nets (base)} & \tc{Our RS-Net} &\\
\midrule[0.7pt]
\multirow{5}{*}{\tabincell{c}{Quantized\\$\,$ResNet18}}
& $224\times224$ & 68.0 / 88.0 & \textbf{68.8 / 88.4 \textsubscript{(+0.8)}} && 64.9 / 86.0 & \textbf{65.8 / 86.4 \textsubscript{(+0.9)}} &\\
& $192\times192$ & 66.4 / 86.9 & \textbf{67.6 / 87.8 \textsubscript{(+1.2)}} && 63.1 / 84.7 & \textbf{64.8 / 85.8 \textsubscript{(+1.7)}} &\\
& $160\times160$ & 64.5 / 85.5 & \textbf{66.0 / 86.5 \textsubscript{(+1.5)}} && 61.1 / 83.3 & \textbf{62.9 / 84.2 \textsubscript{(+1.8)}} &\\
& $128\times128$ & 61.5 / 83.4 & \textbf{63.1 / 84.5 \textsubscript{(+1.6)}} && 58.1 / 80.8 & \textbf{59.3 / 81.9 \textsubscript{(+1.2)}} &\\
& $96\times96$ & 56.3 / 79.4 & \textbf{56.6 / 79.9 \textsubscript{(+0.3)}} && 52.3 / 76.4 & \textbf{52.5 / 76.7 \textsubscript{(+0.2)}} &\\
\midrule[0.7pt]
\multirow{5}{*}{\tabincell{c}{Quantized\\$\,$ResNet50}}
& $224\times224$ &  74.6 / 92.2 & \textbf{76.0 / 92.8 \textsubscript{(+1.4)}} && 72.2 / 90.8 & \textbf{74.0 / 91.5 \textsubscript{(+1.8)}} &\\
& $192\times192$ & 73.5 / 91.3 & \textbf{75.1 / 92.4 \textsubscript{(+1.6)}} && 70.9 / 89.8 & \textbf{73.1 / 91.0 \textsubscript{(+2.2)}} &\\
& $160\times160$ & 71.9 / 90.4 & \textbf{73.8 / 91.6 \textsubscript{(+1.9)}} && 69.0 / 88.5 & \textbf{71.4 / 90.0 \textsubscript{(+2.4)}} &\\
& $128\times128$ &  69.6 / 88.9 & \textbf{71.7 / 90.2 \textsubscript{(+2.1)}} && 66.6 / 86.9 & \textbf{68.9 / 88.3 \textsubscript{(+2.3)}} &\\
& $96\times96$ &  65.5 / 86.0 &\textbf{67.3 / 87.4 \textsubscript{(+1.8)}} && 61.7 / 83.4 & \textbf{63.4 / 84.7 \textsubscript{(+1.7)}} &\\

\bottomrule[1.2pt]
\end{tabular}}
\label{tabs:quan_results}
\vskip-0.3em

\end{table}

\begin{table}[t]
\centering

\caption{Top-1 accuracies (\%) on MobileNetV2. Individual models (I-Nets-w, adjust via network width; I-Nets-r, adjust via resolution) and our RS-Net are trained under the same settings. Results of S-MobileNetV2 (S) \cite{DBLP:conf/iclr/YuYXYH19} and US-MobileNetV2 (US) \cite{DBLP:journals/corr/abs-1903-05134} are from the original papers.}
\vskip-0.1em
\resizebox{108mm}{!}{
\begin{tabular}{p{0.15cm}<{\centering}p{0.9cm}|p{1.2cm}<{\centering}|p{1.4cm}<{\centering}p{1.1cm}<{\centering}p{1.4cm}<{\centering}||p{1.7cm}<{\centering}|p{1.2cm}<{\centering}|p{1.4cm}<{\centering}p{1.1cm}<{\centering}} 
\toprule[1.2pt]
&Width&MAdds&I-Nets-w & S \cite{DBLP:conf/iclr/YuYXYH19} & US \cite{DBLP:journals/corr/abs-1903-05134} & Resolution & MAdds & I-Nets-r & Ours \\
\midrule[0.7pt]
&$1.0\times$ & 301M & 72.1 & 70.5 & 71.5 & $224\times224$ & 301M & 72.1 & \textbf{73.0}\\
&$0.8\times$ & 222M & 69.8 & - & 70.0 & $192\times192$ & 221M & 71.0 & \textbf{72.2}\\
&$0.65\times$ & 161M & 68.0 & - & 68.3 & $160\times160$ & 154M & 69.5 & \textbf{71.1}\\
&$0.5\times$ & 97M & 64.6 & 64.4 & 65.0 & $128\times128$ & 99M & 66.8 & \textbf{68.8}\\
&$0.35\times$ & 59M & 60.1 & 59.7 & 62.2 & $96\times96$ & 56M & 62.6 & \textbf{63.9}\\
\midrule
\multicolumn{3}{c|}{Model Size} & $5\times$ & $1\times$ & $1\times$ & \multicolumn{2}{c|}{Model Size} & $5\times$ &  $1\times$  \\

\bottomrule[1.2pt]
\end{tabular}}
\label{tabs:mflops}
\vskip-0.9em
\end{table}

\subsubsection{Quantization.}
\vskip -1em
We further explore the generalization of our method to more challenging quantization problems, and we apply our method to LQ-Nets \cite{DBLP:conf/eccv/ZhangYYH18}. Experiments are performed under two typical kinds of quantization settings, including the quantization on weights (2$/$32) and the more extremely compressed quantization on both weights and activations (2$/$2). Results of I-Nets and each RS-Net based on quantized ResNets are reported in Table \ref{tabs:quan_results}. Again, each RS-Net outperforms the corresponding I-Nets at all resolutions. For quantization problems, as I-Nets cannot force the quantized parameter values of each individual model to be the same, 2-bit weights in I-Nets are practically stored with more digits than a 4-bit model\footnote{The total bit number of five models with individual 2-bit weights is $\log_2(2^2\times5)\approx4.3$.}, while the RS-Net avoids such issue. For quantization, accuracy gains of RS-Net to ResNet50 are more obvious than those to ResNet18, we conjecture that under compressed conditions, ResNet50 better bridges the network capacity and the augmented data resolutions.

\subsubsection{Switchable Models Comparison.} 
\vskip -1em
Based on MobileNetV2, results in Table \ref{tabs:mflops} indicate that under comparable FLOPs (Multiply-Adds), adjusting image resolutions (see I-Nets-r) achieves higher accuracies than adjusting network widths (see I-Nets-w). For example, adjusting resolution to $96\times96$ brings 2.5\% higher absolute accuracy than adjusting width to $0.35\times$, with even lower FLOPs. Results of S-MobileNetV2 (S) \cite{DBLP:conf/iclr/YuYXYH19} and US-MobileNetV2 (US) \cite{DBLP:journals/corr/abs-1903-05134} are also provided for comparison. As we can see, our RS-Net significantly outperforms S and US at all given FLOPs, achieving 1.5\%$\sim$4.4\% absolute gains. Although both S and US can also adjust the accuracy-efficiency trade-off, they have marginal gains or even accuracy drops (e.g., at width $1.0\times$) compared with their baseline I-Nets-w. Our model shows large gains compared with our baseline I-Nets-r (even they are mostly stronger than I-Nets-w). Note that adjusting resolutions does not conflict with adjusting widths, and both methods can be potentially combined together.

\begin{figure}[t]
\hskip-0.1em
\includegraphics[scale=0.39]{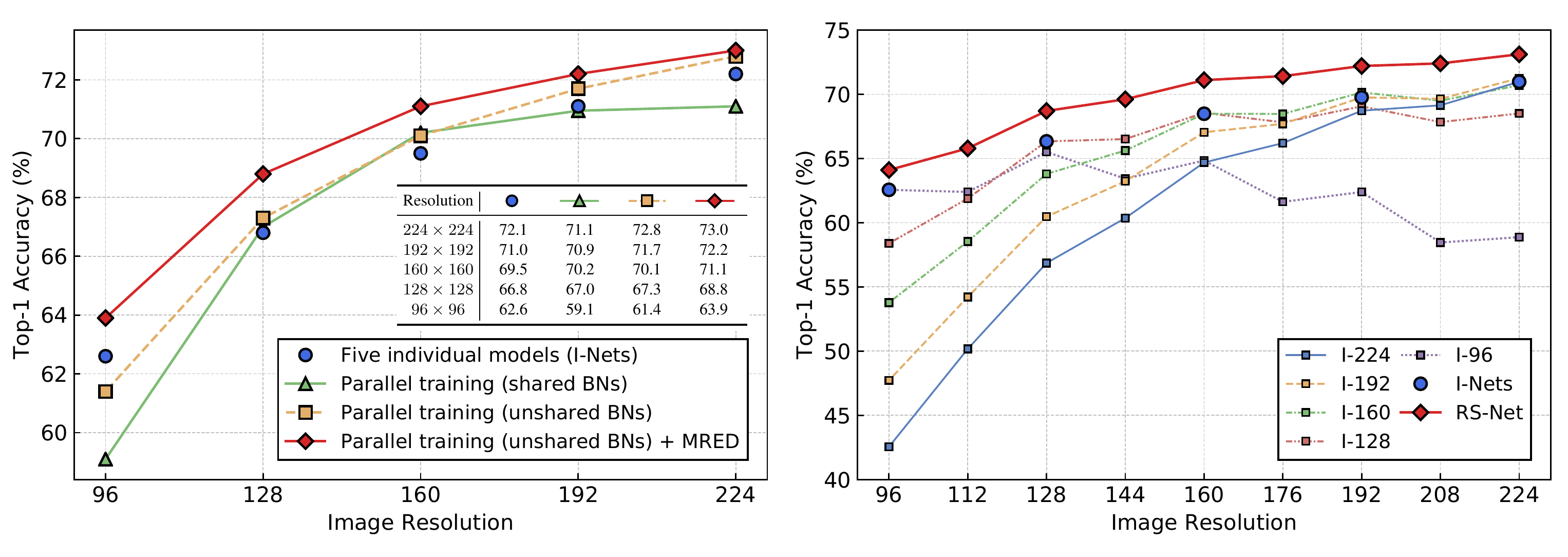}
\vskip -1em
\caption{\textbf{Left:} Comparison of parallel trainings with shared BNs and unshared (private) BNs, based on MobileNetV2. Individual models and Parallel (unshared BNs) + MRED (i.e., RS-Net) are provided for reference. \textbf{Right:} Comparison of our RS-Net and each individual model (from I-$224$ to I-$96$) tested at denser resolutions (the interval is 16), based on ResNet18. Each individual model suffers from serious accuracy drops at other resolutions, but our RS-Net avoids this issue.}
\label{share}
\vskip -0.5em
\end{figure}

\subsection{Ablation Study}

\subsubsection{Importance of Using Private BNs.}
A quick question is that, why not share BNs as well? Based on MobileNetV2, the left panel of Fig. \ref{share} shows that during the parallel training, privatizing BNs achieves higher accuracies than sharing BNs, especially at both the highest resolution (+1.7\%) and the lowest (+2.3\%). When BNs are shared, activation statistics of different image resolutions are averaged, which differ from the real statistics especially at two ends of resolutions.

\subsubsection{Tested at New Resolutions.} 
\vskip -1em
One may concern that if a RS-Net can be tested at a new resolution. In the right panel of Fig. \ref{share}, we test models at different resolutions with a smaller interval, using ResNet18. We follow a simple method to handle a resolution which is not involved in training. Suppose the resolution is sandwiched between two of the five training resolutions which correspond to two groups of BN parameters, we apply a linear interpolation on these two groups and obtain a new group of parameters, which are used for the given resolution. We observe that the RS-Net maintains high accuracies. RS-Net suppresses the serious accuracy drops which exist in every individual model.

\begin{figure}[t]
\centering
\hskip-0.3em\includegraphics[scale=0.36]{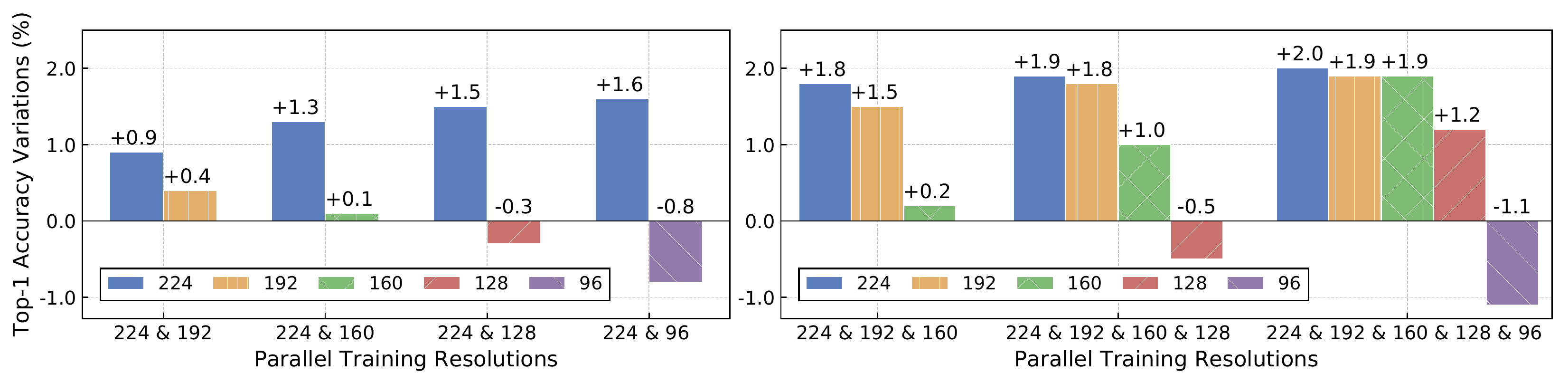}
\vskip-0.6em
\caption{Absolute top-1 accuracy variations (\%) (compared with individual models) of parallel trainings, based on ResNet18. The result of each individual model (from I-$96$ to I-$224$) is used as the baseline. We use single numbers to represent the image resolutions.}
\label{parallel}
\vskip-1em
\end{figure}

\subsubsection{Multi-Resolution Interaction Effects.} 
\vskip-0.7em
This part is for verifying the analysis in Section \ref{sec:discrepancy}, and we do not apply MRED here. Parallel training results of ResNet18 are provided in Fig. \ref{parallel}, which show top-1 accuracy variations compared with each individual model. The left panel of Fig. \ref{parallel} illustrates the parallel training with two resolutions, where all accuracies increase at $224\times224$. As the gap of the two resolutions increases, the accuracy variation at the lower resolution decreases. For example, compared with I-$224$ and I-$96$ respectively, the parallel training with $224\times224$ and $96\times96$ images increases the accuracy at $224\times224$ from 71.0\% to 72.6\%, but decreases the accuracy at $96\times96$ from 62.6\% to 61.8\%. The right panel of Fig. \ref{parallel} illustrates the parallel training with multiple resolutions, where we observe that most accuracies are improved and accuracies at the lowest resolutions may decrease. These results verify our analysis.

\subsubsection{Verification of Ensemble Distillation.} 
\vskip-0.7em
\label{veri_kd}
In Section \ref{mred}, we define two versions of MRED. The vanilla-version has only the distillation paths starting from the ensemble prediction $\bm{p}_0$ toward all the other predictions, while the full-version has additional paths from predications at high resolutions toward predictions at lower resolutions. In Table \ref{tabs:kd_results}, we compare the performance of the two versions as well as two other variants that omit the distillations from $\bm{p}_0$, based on ResNet18. Results indicate that all kinds of the proposed distillations are indispensable. Besides, we also emphasize in Section \ref{mred} that for training a RS-Net, each image should be randomly cropped only once (called single-crop) and then resized to multiple resolutions (called multi-resolution). Results in Table \ref{tabs:kd_crop_results} indicate that using multi-crop (applying a random crop individually for each resolution) will weaken the benefits of MRED, as accuracies at low resolutions are lower compared with using single-crop. We also verify the importance of multi-resolution by replacing the multi-resolution setting by five identical resolutions, called single-resolution. As resolutions are identical, an individual experiment is needed for each resolution. Results indicate that applying ensemble distillation to predictions of different crops has very limited benefits compared with applying it to predictions of different resolutions.

\begin{table}[t]
\centering

\caption{Top-1 accuracies (\%) of different kinds of distillations and their accuracy variations compared with the full-version MRED, based on ResNet18. Definitions of $\bm{p}_s, s\in\{1,2,\cdots,S\}$ are referred to Section \ref{mred}. The prediction $\bm{p}_1$ corresponds to the largest resolution $224\times224$.}

\vskip-0.3em
\resizebox{100mm}{!}{
\begin{tabular}{p{1.7cm}<{\centering}|p{2.2cm}<{\centering}p{2.2cm}<{\centering}p{2.7cm}<{\centering}p{2.5cm}<{\centering}} 
\toprule[1.2pt]
Resolution & \tabincell{c}{Full-version\\MRED (base)} & \tabincell{c}{Vanilla-version\\MRED} & \tabincell{c}{Only distillations\\ from $\bm{p}_1,\bm{p}_2,\cdots,\bm{p}_{S-1}$}   & \tabincell{c}{Only distillations\\ from $\bm{p}_1$}  \\
\midrule[0.7pt]
 $224\times224$ & 73.1 & 73.0 (-0.1) & 72.3 (-0.8)  & 72.2 {(-0.9)}\\
 $192\times192$ &  72.2 & 72.1 {(-0.1)} & 71.2 {(-1.0)} & 71.7 {(-0.5)}\\
 $160\times160$ & 71.1 & 70.8 {(-0.3)} & 70.2 {(-0.9)} & 70.5 {(-0.6)}\\
 $128\times128$ &  68.7 & 68.3 {(-0.4)} & 68.1 {(-0.6)} & 68.1 {(-0.6)}\\
 $96\times96$ & 64.1 & 63.7 {(-0.4)} & 63.5 {(-0.6)} & 63.1 {(-1.0)} \\

\bottomrule[1.2pt]

\end{tabular}}
\label{tabs:kd_results}
\vskip-0.4em
\end{table}

\begin{table}[t]
\centering

\caption{Top-1 accuracies (\%) comparison to verify the importance of using multi-resolution and single-crop, based on ResNet18. Multi-crop refers to applying the random crop individually for each resolution. Single-resolution refers to using five identical resolutions in a RS-Net instead of the original multi-resolution setting.}
\vskip-0.3em
\resizebox{100mm}{!}{
\begin{tabular}{p{1.7cm}<{\centering}|p{2.4cm}<{\centering}p{2.4cm}<{\centering}p{2.4cm}<{\centering}p{2.4cm}<{\centering}} 
\toprule[1.2pt]
Resolution & \tabincell{c}{Multi-resolution\\Single-crop (base)} & \tabincell{c}{Multi-resolution\\Multi-crop} & \tabincell{c}{Single-resolution\\Multi-crop} & \tabincell{c}{Single-resolution\\Single-crop}\\
\midrule[0.7pt]
 $224\times224$ & 73.1 & 73.1 (-0.0)  & 72.1 (-1.0) & 71.0 {(-2.1)}\\
 $192\times192$ &  72.2 & 72.0 {(-0.2)} & 71.1 (-1.1) & 69.8 (-2.4)\\
 $160\times160$ & 71.1 & 70.7 {(-0.4)} & 69.6 (-1.5) & 68.5 (-2.6)\\
 $128\times128$ &  68.7 & 68.0 {(-0.7)} & 67.6 (-1.1) & 66.3 (-2.4)\\
 $96\times96$ & 64.1 & 62.8 {(-1.3)} & 63.3 (-0.8) & 62.6 (-1.5)\\
\midrule
{Model Size} & $1\times$ & $1\times$ & $5\times$ &  $5\times$  \\

\bottomrule[1.2pt]

\end{tabular}}

\label{tabs:kd_crop_results}
\vskip-1.2em
\end{table}

\section{Conclusions}
\vskip -0.39em
We introduce a general method to train a CNN which can switch the resolution during inference, and thus its running speed can be selected to fit different computational constraints. Specifically, we propose the parallel training to handle multi-resolution images, using shared parameters and private BNs. We analyze the interaction effects of resolutions from the perspective of train-test discrepancy. And we propose to learn an ensemble distillation based on the same image instances with different resolutions, which improves accuracies to a large extent.

\section*{Acknowledgement}
\vskip -0.35em
This work is  jointly supported by the National Science Foundation of China (NSFC) and the German Research Foundation (DFG) in project Cross Modal Learning, NSFC 61621136008/DFG TRR-169. We thank Aojun Zhou for the insightful discussions.

\clearpage

\section*{\LARGE Appendix}
\appendix
\vskip0.6em

\section{Details of the Ensemble} As the importance scores $\bm{\alpha}$ for ensemble are essential to the MRED, we study their values w.r.t. different resolutions. We observe that  the learned values of $\bm{\alpha}$ stay almost unchanged in the last three epochs. The final results w.r.t. the resolutions from $224\times224$ to $96\times96$ are $\{0.37,0.29,0.20,0.12,0.02\}$ for ResNet18, $\{0.32,0.30,0.23,0.13,0.02\}$ for ResNet50, and $\{0.41,0.30,0.19,0.09,0.01\}$ for MobileNetV2. In each network, the score w.r.t. the resolution $224\times224$ has the largest ratio, and the ratio decreases with the decrease of resolutions.

\section{Visualization of BNs.} 
\begin{figure}[h]
\centering
\includegraphics[scale=0.2]{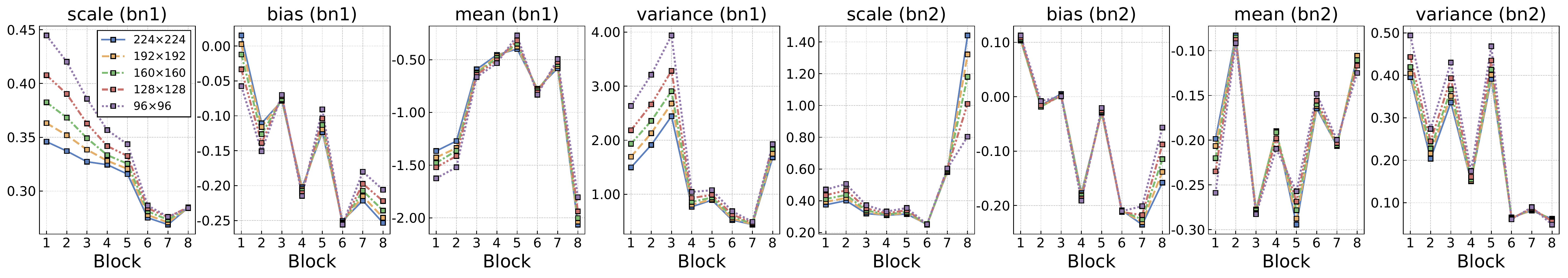}
\vskip-0.5em
\caption{BN parameters and statistics in ResNet18 blocks. The first layer of each block is shown in the left four sub-figures, and the second is shown in the right four sub-figures.}
\vskip-2em
\label{pic_resnet18_bn}
\end{figure}

Fig. \ref{pic_resnet18_bn} visualizes BN parameters, including scale $\bm{\gamma}$, bias $\bm{\beta}$, mean $\bm{\mu}$ and variance $\bm{\sigma}$, in a parallel trained model on ResNet18. There are eight blocks and each has two Conv layers. We plot the channel-wise means of BN parameters of every first layer in the left four sub-figures and of every second layer in the right four sub-figures. We observe that BN parameters are likely to be arranged in the ascending order or the descending order of image resolutions.
\vskip-0.5em

\section{Extension to Semantic Segmentation}

\begin{figure}[t]
\centering
\hskip-0.1em
\includegraphics[scale=0.272]{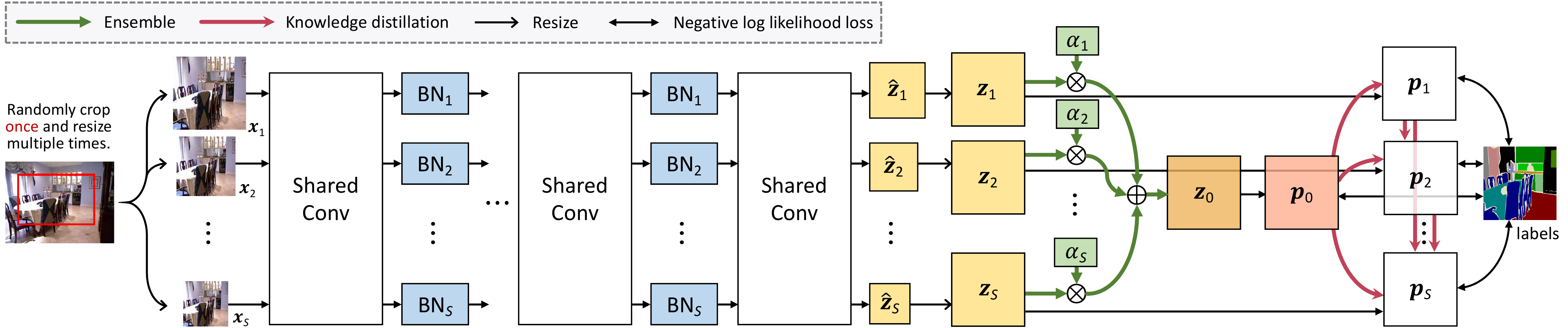}

\caption{Framework of training a RS-Net for semantic segmentation. Logits outputted by the last Conv layer are denoted as $\hat{\bm{z}}_1,\hat{\bm{z}}_2,\cdots,\hat{\bm{z}}_S$. We resize these logits to the same resolution of $\bm{x}_1$, which is the largest input resolution. We denote the resized logits as $\bm{z}_1,\bm{z}_2,\cdots,\bm{z}_S$. The ensemble logit $\bm{z}_0$ is learned as a weighted mean of the resized logits. During testing, each logit $\hat{\bm{z}}_s,s\in\{1,2,\cdots,S\}$ is uniformly resized to the original image resolution for evaluation.}
\label{framework_seg}
\vskip-1em
\end{figure}

Besides the experiments described in the main paper, we also apply our method to semantic segmentation to further validate its generalization ability to handle other visual recognition tasks beyond classification.
We choose RefineNet \cite{DBLP:conf/cvpr/LinMSR17}, a typical semantic segmentation model which achieves state-of-the-art results on dataset NYUDv2 \cite{Silberman:ECCV12}. Following the original setting in \cite{DBLP:conf/cvpr/LinMSR17}, we use ResNet101 as the backbone network. A schematic framework for training a RS-Net for semantic segmentation is illustrated in Figure \ref{framework_seg}. During training, each logit outputted by the last Conv layer, denoted as $\hat{\bm{z}}_s,s\in\{1,2,\cdots,S\}$, does not has the same resolution with its corresponding input $\bm{x}_s$. For example, if we choose the multi-resolution setting as $\mathbb{S}=\{352\times352,224\times224,96\times96\}$, resolutions of $\hat{\bm{z}}_1,\hat{\bm{z}}_2,\hat{\bm{z}}_3$ will be $88\times88,56\times56,24\times24$ respectively. We uniformly resize $\hat{\bm{z}}_s$ to the largest input resolution (for this example is $352\times352$) before the ensemble distillation process and calculating losses with labels. We do not use left-right flips or the multi-scale technique during testing for additional performance promotion, and each logit $\hat{\bm{z}}_s,s\in\{1,2,\cdots,S\}$ is uniformly resized to the original image resolution before calculating evaluation metrics.

Results on NYUDv2 are shown in Table \ref{tabs:seg_results}. Following \cite{DBLP:conf/cvpr/LinMSR17}, we train on RGB images with 40 classes, using the standard training and testing split with 795 and 654 images respectively. These results verify that our method can be applied to the semantic segmentation task, maintaining the resolution switchable ability while simultaneously improves performance. As far as we know, this is the first resolution switching attempt for semantic segmentation, realizing a selectable inference speed which is beneficial to efficient runtime model deployments. As we can see in Table \ref{tabs:seg_results}, RS-Net especially achieves performance gains over I-Nets at low resolutions, e.g., with a significant IoU gain 14.0 at $96\times96$.

\begin{figure}[t]
\centering
\hskip-0.2em
\includegraphics[scale=0.51]{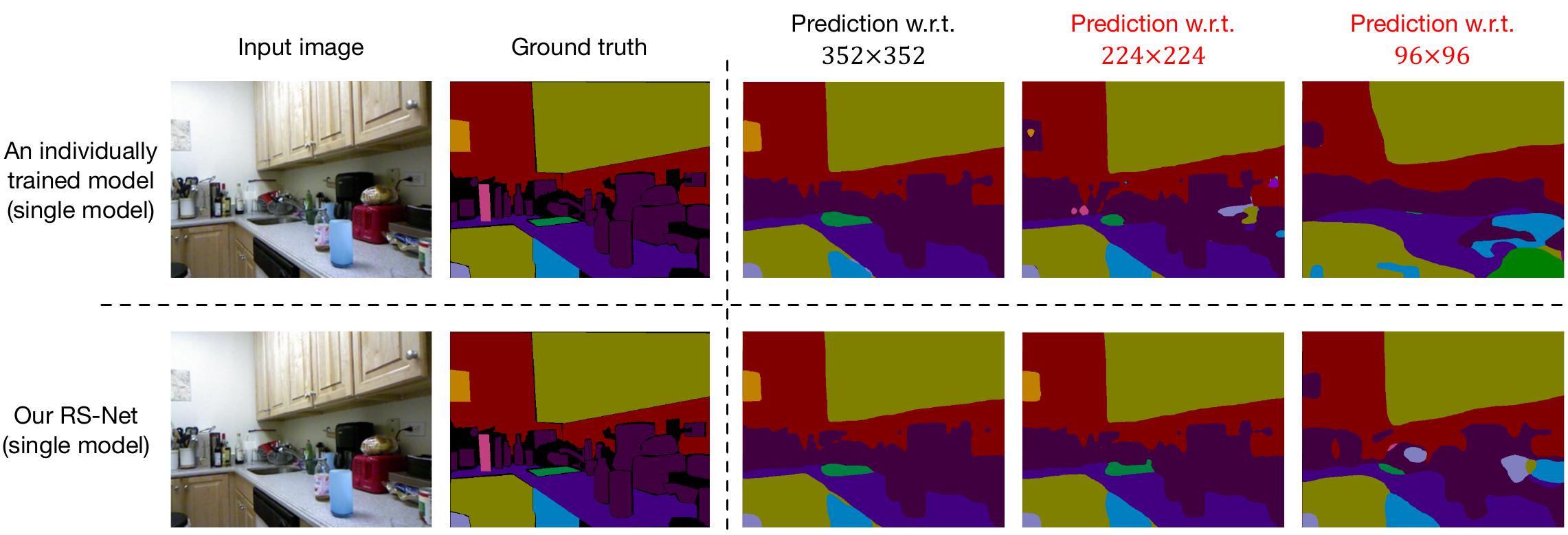}

\caption{Performance comparison of an individual model and our RS-Net. The rightmost four sub-figures (with titles marked in red) verify that our model can better maintain the performance when input resolution at inference is downsized for the sake of saving inference time.}
\label{pic_seg}
\end{figure}

In Fig. \ref{pic_seg}, we compare our RS-Net with an individual model which is trained at $352\times352$. We evaluate performance at three resolutions during inference, for proving that our model has better robustness against various resolutions. Predictions w.r.t. $224\times224$ and $96\times96$ indicate that downsizing input resolution leads to quick performance drops for an individual model. In contrast, our RS-Net has milder performance drops toward downsizing the resolution.

\begin{table}[t]
\centering

\caption{Results comparison for semantic segmentation based on RefineNet with ResNet101 as the backbone. We report pixel accuracies (\%), mean accuracies (\%) and IoU of individual models (I-Nets) and our RS-Net. Note that no left-right flips or multi-scale testing is performed. All experiments use the same data pre-processing methods and training settings. }
\resizebox{115mm}{!}{
\begin{tabular}{p{2.2cm}<{\centering}|p{1.5cm}<{\centering}p{1.3cm}<{\centering}p{1cm}<{\centering}|p{1.9cm}<{\centering}p{1.9cm}<{\centering}p{1.9cm}<{\centering}} 

\toprule[1pt]
 \multirow{2}*{Resolution} & \multicolumn{3}{c|}{I-Nets (base)}& \multicolumn{3}{c}{Our RS-Net} \\
&\tc{Pixel Acc.} & \tc{Mean Acc.} & \tcc{IoU} & \tc{Pixel Acc.} & \tc{Mean Acc.} & \tc{IoU}\\
\midrule[0.7pt]
 $352\times352$ & 72.3 & 56.7 & 43.9 & \textbf{72.3 \textsubscript{(+0.0)}} & \textbf{57.0 \textsubscript{(+0.3)}}& \textbf{44.1 \textsubscript{(+0.2)}}\\
 $224\times224$ & 69.6 & 52.2 & 40.5 & \textbf{71.4 \textsubscript{(+1.8)}} & \textbf{54.6 \textsubscript{(+2.4)}}& \textbf{42.6 \textsubscript{(+2.1)}}\\
 $96\times96$ & 50.4 & 26.5 &18.1 & \textbf{ 63.0 \textsubscript{(+12.6)}} & \textbf{ 43.1 \textsubscript{(+16.6)}}& \textbf{ 32.1 \textsubscript{(+14.0)}}\\
\midrule
{Total Params} & \multicolumn{3}{c|}{$118.20\text{M}\times3=354.60\text{M}$} & \multicolumn{3}{c}{$118.31\text{M}$}   \\

\bottomrule[1pt]
\end{tabular}}
\label{tabs:seg_results}
\vskip-2em
\end{table}

\begin{table}[t]
\centering

\caption{Top-1 accuracies (\%) comparison at different testing resolutions. Our model has better performance especially at low resolutions, which means being able to achieve better accuracy-efficiency trade-offs at runtime. Note that the accuracy at $224\times224$ of RS-Net has already surpassed all results of FixRes.}

\resizebox{110mm}{!}{
\begin{tabular}{p{3.3cm}<{\centering}|p{0.1cm}p{1.2cm}<{\centering}p{1.2cm}<{\centering}p{1.2cm}<{\centering}p{1.2cm}<{\centering}p{1.2cm}<{\centering}p{1.2cm}<{\centering}p{1.2cm}<{\centering}p{1.2cm}}
\toprule[1pt]
Model $\setminus$ Resolution && \tc{$64$} & \tc{$128$} & \tc{$224$} &\tc{$288$} & \tc{$352$}  & \tc{$384$} & \tc{$448$} \\
\midrule[0.7pt]
 FixRes \cite{DBLP:journals/corr/abs-1906-06423} && 41.7 & 67.7 & 77.1  & 78.5 & \textbf{78.9} & \textbf{79.0} & \textbf{78.4}\\
 Our RS-Net &&\textbf{61.1} & \textbf{76.3} & \textbf{79.3} & \textbf{79.2} & 78.1 & 77.4 & 75.8 \\
 \midrule[0.7pt]
Multiply-Adds && \tc{338M} & \tc{1.35G} & \tc{4.14G} & \tc{6.84G} & \tc{10.22G} & \tc{12.17G} & \tc{16.56G} \\

\bottomrule[1pt]

\end{tabular}}
\label{tabs:fixres}
\vskip-0.3em
\end{table}

\section{Comparison with FixRes}
\vskip-0.3em
Although FixRes \cite{DBLP:journals/corr/abs-1906-06423} and our work have both considered resolution adaptation, they are different in motivation and design. FixRes focuses on improving accuracy by operating models at much higher resolution at test time, relying on manual fine-tuning for adaptation and test-time augmentations. However, our method focuses on efficient and flexible resolution adaptation at test time without additional latency such as fine-tuning. The accuracy comparison is possible as we both consider experiments on ResNet-50. In Fig.5 of FixRes and Table 5 of its supplementary material, for a ResNet-50 model trained with $224\times224$ images, the top-1 accuracy drops 9.4\% from $224\times224$ (77.1\%) to $128\times128$ (67.7\%), while ours merely drops 3.0\% from $224\times224$ (79.3\%) to $128\times128$ (76.3\%) (Table \ref{tab_proportion}). Therefore, our method can better suppress the accuracy drop when input image resolution is downsized, which is beneficial to the model deployment in a resource-constrained platform. 
A more detailed comparison is shown in Table \ref{tabs:fixres}, which indicates that our model has much better performance at low resolutions, saving a large amount of FLOPs but even achieving higher performance. For example, our accuracy at $224\times224$ surpasses the top accuracy of FixRes at $384\times384$, needing only 34\% FLOPs. We conjecture that under the permission of training resources, our RS-Net has the potential to achieve better performance by adding larger resolutions (e.g. $384\times384$ or larger) for training.

\section{Discrepancy and Interaction Effects}
We conduct an additional contrast experiment for verifying our analysis in Section \ref{sec:parallel_training}, where we propose that the multi-resolution interaction effects are highly correlated with the train-test discrepancy, which is a kind of distribution shift caused by different data pre-processing methods during training and testing. As a conclusion of our analysis,  on account of the multi-resolution parallel training, accuracies at higher resolutions tend to be further improved, but the accuracy at the low resolution tends to be reduced. In this part, we try to reduce the train-test discrepancy and observe if such interaction effects are weakened.

\begin{figure}[t]
\centering
\hskip-0.4em
\includegraphics[scale=0.35]{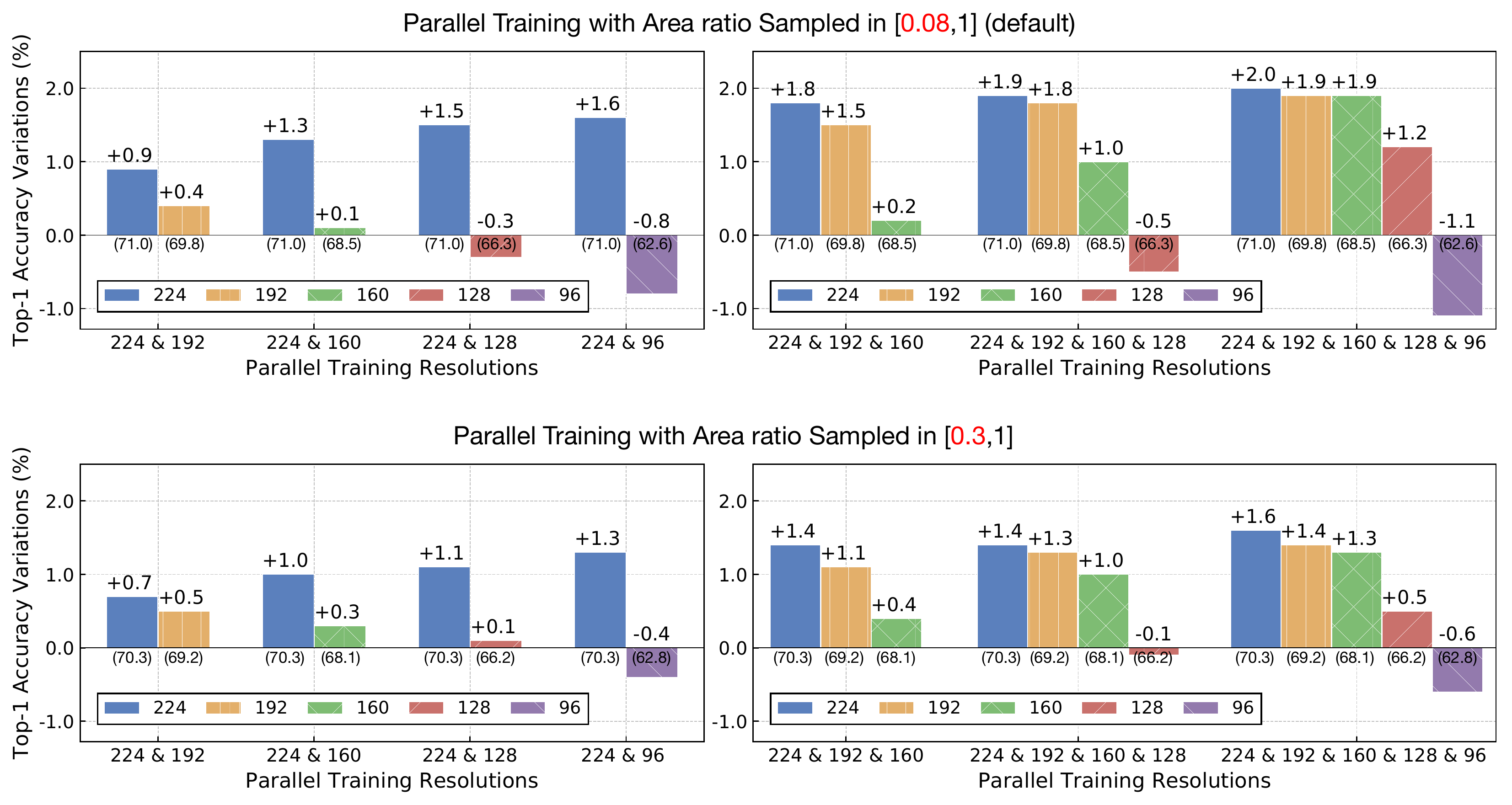}
\vskip-0.8em
\caption{Absolute top-1 accuracy variations (\%) (compared with individual models) of parallel trainings, based on ResNet18, with two settings of the area ratio. The top-1 accuracy (\%) of each individual model (from I-$96$ to I-$224$) is written in the bracket, which is used as the baseline. We use single numbers to represent the image resolutions.}
\label{parallel}
\vskip-0.5em
\end{figure}

The concept of the train-test discrepancy itself is revealed by \cite{DBLP:journals/corr/abs-1906-06423}. We first re-explain this discrepancy, based on our experiment setting as a specific example.
In Section 4.1 of our paper, we mention that during training, we randomly crop the data for augmentation with an area ratio\footnote{The area ratio means the ratio of the cropped image area to the original image area.} uniformly sampled in $[0.08, 1.0]$, which is a standard setting following \cite{DBLP:conf/cvpr/HuangLMW17,DBLP:conf/cvpr/SzegedyLJSRAEVR15,DBLP:journals/corr/abs-1906-06423}. Therefore the expectation of area ratio for training is $(0.08+1)/2=0.54$. During testing, we first resize images to the target resolution divided by $0.875$ (following \cite{DBLP:journals/corr/abs-1908-08986,DBLP:journals/corr/abs-1908-03888}), and then crop the central regions with the target resolution. Therefore the expectation of area ratio for testing is $0.875^2\approx0.77$, which is larger than during training. As a larger crop means a smaller apparent object size, so on average, the apparent object size in testing is smaller than in training, which is the so-called train-test discrepancy \cite{DBLP:journals/corr/abs-1906-06423}. Note that the parameters $[0.08, 1.0]$ and $0.875$ are not always adopted by all image recognition works, but the train-test discrepancy typically exists (in different degrees) \cite{DBLP:journals/corr/abs-1906-06423}.

We alleviate the discrepancy by modifying $[0.08, 1.0]$ to $[0.3, 1.0]$, because the expectation of area ratio for training becomes $(0.3+1)/2=0.65$, which is closer to $0.77$ in testing. Results of parallel training (without MRED) are illustrated in Fig. \ref{parallel}, including top-1 accuracy variations over each individual model. We can see that by alleviating the discrepancy, interaction effects are weakened, as accuracy gains at high resolutions and accuracy drops at the lowest resolution are both alleviated. Besides, in Fig. \ref{parallel}, we also provide the accuracy of each individual model (see each number in the bracket). We observe that sampling the area ratio in $[0.08, 1.0]$ has better overall performance than $[0.3, 1.0]$, which also indicates why $[0.08, 1.0]$ is a more popular choice for training on ImageNet.

\bibliographystyle{splncs04}
\bibliography{egbib}

\end{document}